\documentclass[10pt,twocolumn,letterpaper]{article}

\usepackage{cvpr}              %

\usepackage{graphicx}
\usepackage{amsmath}
\usepackage{amssymb}
\usepackage{booktabs}
\usepackage{color}
\usepackage{multirow}
\usepackage[pagebackref,breaklinks,colorlinks]{hyperref}

\usepackage[capitalize]{cleveref}
\crefname{section}{Sec.}{Secs.}
\Crefname{section}{Section}{Sections}
\Crefname{table}{Table}{Tables}
\crefname{table}{Tab.}{Tabs.}
\usepackage{textcomp}

\usepackage{overpic}
\usepackage{enumitem} %
\usepackage{overpic} %
\usepackage{color}
\usepackage{booktabs}
\usepackage{gensymb}
\usepackage{colortbl}
\usepackage{pifont}%
\usepackage{bm}
\definecolor{turquoise}{cmyk}{0.65,0,0.1,0.3}
\definecolor{purple}{rgb}{0.65,0,0.65}
\definecolor{dark_green}{rgb}{0, 0.5, 0}
\definecolor{orange}{rgb}{0.8, 0.6, 0.2}
\definecolor{red}{rgb}{0.8, 0.2, 0.2}
\definecolor{darkred}{rgb}{0.6, 0.1, 0.05}
\definecolor{blueish}{rgb}{0.0, 0.3, .6}
\definecolor{light_gray}{rgb}{0.7, 0.7, .7}
\definecolor{pink}{rgb}{1, 0, 1}
\definecolor{greyblue}{rgb}{0.25, 0.25, 1}

\crefname{section}{\S}{\S\S}
\crefname{subsection}{\S}{\S\S}
\crefname{equation}{\text{Eq}}{\text{Eq}}
\crefname{definition}{\text{Dfn.}}{\text{Dfn.}}

\usepackage{blindtext}

\renewcommand{\paragraph}[1]{\vspace{1em}\noindent\textbf{#1}.}

\newcommand{\beginsupplement}{%
        \setcounter{table}{0}
        \renewcommand{\thetable}{A\arabic{table}}%
        \setcounter{figure}{0}
        \renewcommand{\thefigure}{A\arabic{figure}}%
     }

\makeatletter
\DeclareRobustCommand\onedot{\futurelet\@let@token\@onedot}
\def\@onedot{\ifx\@let@token.\else.\null\fi\xspace}

\def\eg{\emph{e.g}\onedot} 
\def\ie{\emph{i.e}\onedot}

\def\etal{\emph{et al}\onedot}
\makeatother

\newcommand{\static}{\mathrm{static}}
\newcommand{\loss}{\mathcal{L}}

\newcommand{\T}{\mathbf{T}}

\newcommand{\Feature}{\mathbf{F}}
\newcommand{\feature}{\mathbf{f}}

\newcommand{\point}{\mathbf{x}}

\newcommand{\cat}{\mathrm{cat}}

\newcommand{\PC}{\mathbf{X}}
\newcommand{\PCset}{\mathcal{X}}
\newcommand{\ImageSet}{\mathcal{I}}
\newcommand{\Image}{\mathbf{I}}
\newcommand{\flow}{\mathrm{flow}}
\newcommand{\depth}{\mathrm{depth}}
\newcommand{\cycle}{\mathrm{cycle}}
\newcommand{\K}{\mathbf{K}}
\newcommand{\Point}{\mathbf{P}}

\newcommand{\photo}{\mathrm{photo}}
\newcommand{\smooth}{\mathrm{smooth}}
\newcommand{\OF}{\mathbf{O}}
\newcommand{\deflow}{\textsc{DeFlow}}
\newcommand{\Mask}{\mathcal{M}}

\definecolor{lossred}{rgb}{1.0, 0.01, 0.24}
\definecolor{lossgreen}{rgb}{0.55, 0.71, 0.0}
\definecolor{lossblue}{rgb}{0.0, 0.44, 1.0}
\definecolor{lossyellow}{rgb}{1.0, 0.66, 0.07}
\definecolor{losspurple}{rgb}{0.76, 0.33, 0.76}
\definecolor{tab10orange}{rgb}{1.0, 0.7, 0.0}

\def\cvprPaperID{2} %
\def\confName{CVPR}
\def\confYear{2023}

\begin{document}

\title{\deflow: Self-supervised 3D Motion Estimation of Debris Flow}

\author{ Liyuan Zhu\textsuperscript{1} \quad
Yuru Jia\textsuperscript{1} \quad
Shengyu Huang\textsuperscript{1} \quad
Nicholas Meyer\textsuperscript{1} \\
 Andreas Wieser\textsuperscript{1} \quad
Konrad Schindler\textsuperscript{1} \quad
Jordan Aaron\textsuperscript{2,3} 
\quad
\vspace{5px}
\\
\textsuperscript{1}{\normalsize Institute of Geodesy and Photogrammetry, ETH Zurich} \\
\textsuperscript{2}{\normalsize Swiss Federal Institute for Forest, Snow and Landscape Research WSL} \\
\textsuperscript{3}{\normalsize Geological Institute, ETH Zurich} \\
\vspace{-15px}
}
\maketitle
\begin{abstract}
Existing work on scene flow estimation focuses on autonomous driving and mobile robotics, while automated solutions are lacking for motion in nature, such as that exhibited by debris flows. We propose \deflow, a model for 3D motion estimation of debris flows, together with a newly captured dataset. We adopt a novel multi-level sensor fusion architecture and self-supervision to incorporate the inductive biases of the scene. We further adopt a multi-frame temporal processing module to enable flow speed estimation over time. Our model achieves state-of-the-art optical flow and depth estimation on our dataset, and fully automates the motion estimation for debris flows.
Source code and dataset are available at \href{https://www.zhuliyuan.net/deflow}{project page}. 
\end{abstract}

\section{Introduction}
\label{sec:intro}

\begin{figure}[t]
    \centering
    \vspace{0mm}
    \includegraphics[width=0.95\linewidth]{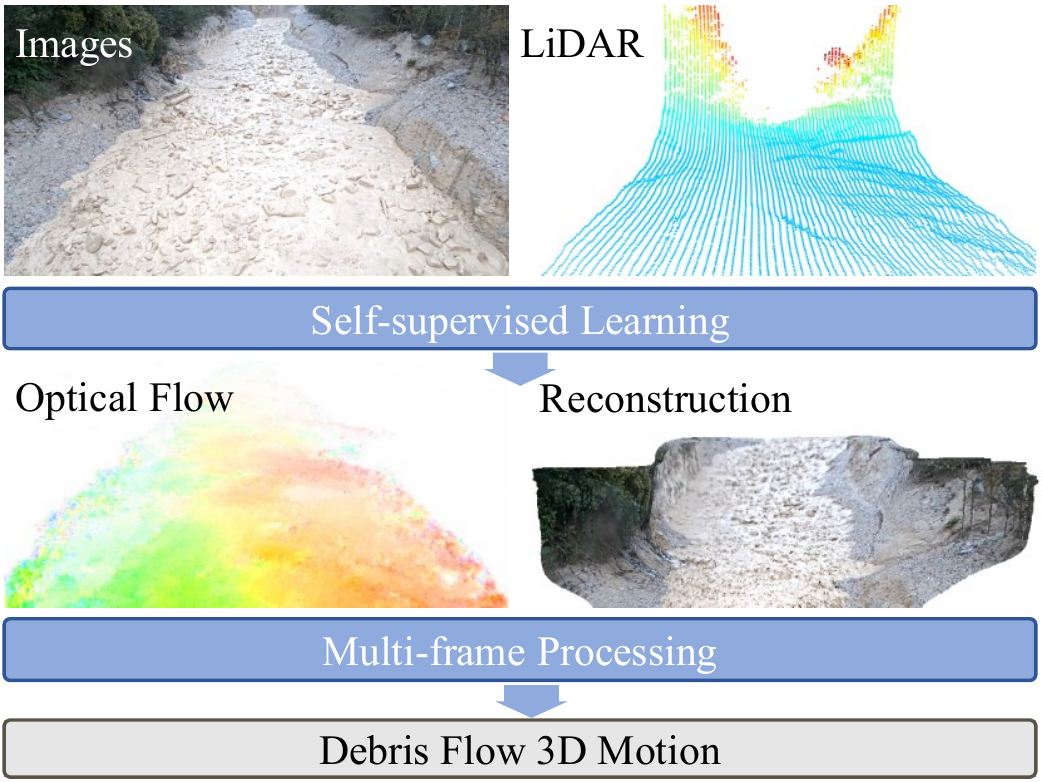}
    \vspace{-2mm}
    \caption{\textbf{Debris flow estimation from images and sparse LiDAR points}. Our self-supervised method automates the estimation of dense scene depth and flow. Additionally, a multi-frame temporal processing module recovers flow velocity profiles over time. 
    }
    \vspace{-6mm}
   \label{fig:teaser}
\end{figure}

Cameras and LiDAR sensors are complementary for 3D scene understanding, consequently many algorithms designed for autonomous driving and mobile robotics use these two modalities. However, little attention has been paid to natural scenes. There is potential to use these sensors to analyse natural processes, and debris flows are particularly amenable to this~\cite{hutter1994debris,holst2021increasing,jordan2023debris}. Debris flows are extremely rapid flows of soil, water and woody debris that can reach velocities in excess of 5 m/s, and travel for multiple kilometers ~\cite{hungr2014varnes}. They are a considerable hazard in mountainous regions and cause costly disasters every year~\cite{petley2012global}. Reducing the impact of debris-flows requires a better understanding of the fundamental mechanisms which drive motion, and progress has been previously limited by a lack of high temporal and spatial resolution observations of debris-flow velocity.
In the present work, we analyse a debris-flow dataset captured with a camera-LiDAR setup (see \cref{fig:teaser}) and estimate dense surface velocity fields. 

Two broad categories of methods can be considered when analyzing natural debris-flows using camera and LiDAR sensors: flow estimation techniques or 3D scene flow. Recent work on fluid flow estimation~\cite{lasinger20203d,zhang2022learning,lasinger2017volumetric} has focused on embedding physical constraints into  Particle Image Velocity (PIV) solutions.  The fluid datasets used are often either generated by numerical simulations, or collected under well-constrained laboratory conditions, which makes it difficult to generalize to real-world scenarios. 3D scene flow estimation relies heavily on the rigid body motion prior~\cite{PointFlowNet_2019_CVPR,gojcic2021weakly,huang2022dynamic,vogel2013piecewise}, and point-wise correspondence ~\cite{gojcic2021weakly,huang2022dynamic,li2021neural}.  These methods are not well-suited to debris flows, as the bulk motion is non-rigid, although locally rigid motion occurs when boulders and woody debris are present on the flow surface~\cite{aaron2023df_jornal}.
From the above observations, we find that existing methods are hampered by \emph{(i)} inapplicable physical constraints \emph{(ii)} over-reliance on rigid motion priors and \emph{(iii)} an inability to generalize to our hard-to-annotate debris-flow dataset.

To overcome the above shortcomings, we propose a method that is tailored to debris-flow monitoring. With no direct supervision from manual annotation, we introduce a self-supervised optical flow branch~\cite{sun2018pwc,hur2020self} into our network,
which exploits the informative and easily trackable pixel features in optical images. Since optical flow estimations only provide 2D pixel-wise motion, we incorporate a depth estimation branch to densify sparse depth observations from LiDAR and lift the optical flow to 3D scene flow. In contrast to \cite{huang2022dynamic,gojcic2021weakly}, we take into account the unique properties of debris flows and of the monitoring system, namely \emph{(i)} the absence of ego-motion and \emph{(ii)} the presence of a continuous fluid surface without motion-induced occlusion. Through experimentation, we show that the optical flow task and the depth estimation task mutually reinforce each other and that our method's performance is further improved by the inclusion of suitable inductive biases and warping-based multi-frame temporal smoothing. Additionally, our method's lightweight nature makes it easy to train from scratch to analyze new debris flow events. We apply our method to derive dense flow velocity fields of a debris-flow event, at the highest spatial resolution obtained so far for a natural debris flow. Finally, we release our code and debris-flow dataset to the community, as a benchmark for understanding motion in nature.

\section{Related work}
\label{sec:related}

\paragraph{Scene flow from RGB images}
Recent studies have explored a variety of methods to estimate depth and motion from RGB images. Brickwedde \etal~\cite{Brickwedde_2019_ICCV} propose a probabilistic depth estimation network and compute the motion of the scene by combining multi-view geometry and single-view depth information. Yang \etal~\cite{yang2020upgrading} introduce optical expansion and propose a network that learns to expand optical flow and estimate motion in depth.
Hur \etal~\cite{hur2020self} jointly estimate depth and 3D scene flow from monocular images by adopting self-supervised learning.

However, \cite{vogel20113d,Brickwedde_2019_ICCV} heavily rely on a rigid motion prior for moving vehicles, which does not apply to debris flows. \cite{yang2020upgrading,hur2020self} suffer from overfitting to specific camera intrinsics. Their networks are trained on KITTI~\cite{geiger2013vision} with known camera intrinsics, but encounter problems when applied to other datasets, \eg, nuScenes~\cite{caesar2020nuscenes} or our debris-flow data. 

\paragraph{Camera-LiDAR fusion} An intuitive way to disambiguate depth scaling is to incorporate direct 3D measurements into image-based pipelines. Many works have thus focused on sensor fusion for scene flow estimation. LiDAR-Flow~\cite{battrawy2019lidar} fuses the LiDAR with a stereo camera to boost the cost computation in the matching process for both stereo and optical flow. Subsequent work like DeepLiDARFlow~\cite{rishav2020deeplidarflow} encapsulates the solution into an end-to-end deep neural network that regresses scene flow from consecutive LiDAR scans and monocular images. CamLiFlow~\cite{liu2022camliflow} mitigates the sensor fusion problem with a multi-stage fusion framework and estimates point-wise motion for sparse LiDAR point clouds. 

\paragraph{Multi-frame flow processing} 
Two-frame methods~\cite{yang2020upgrading,hur2020self,Brickwedde_2019_ICCV,rishav2020deeplidarflow,liu2022camliflow} do not account for the temporal continuity across multiple consecutive frames. In contrast, \cite{vogel2014view} enforces consistency for both spatial and temporal neighbors in a sliding temporal window, \cite{taniai2017fast} leverages multi-frame consistency to detect moving object regions, and \cite{hur2021self} introduces a convolutional LSTM on top of \cite{hur2020self} to propagate the hidden states via forward warping, but requires stereo images as supervision. Huang \etal~\cite{huang2022dynamic} align multi-frame 3D scans in a common reference frame, accumulate 3D points on individual objects, and obtain a better decomposition of the dynamic scene. Wang \etal~\cite{wang2022neural} encode the spatial-temporal information into a neural trajectory prior. These previous works exploit the geometric layout and rigidity of autonomous driving scenes, but do not generalize well to debris flows. Our work takes temporal information into account and estimates the flow speed across the entire event by warping-based temporal smoothing, which is a stable and efficient choice for debris flows in the absence of strong supervision signals, \eg, stereo images or ground truth scene flow.

\paragraph{Debris flow estimation}
Debris-flow hazard is largely governed by flow depths and velocities. These parameters have only been measured in a few locations worldwide, often using measurement techniques that have poor spatial and/or temporal resolution~\cite{hurlimann2019debris}. Unlike these previous studies, Aaron \etal~\cite{jordan2023debris,Spielmann2023debris,aaron2023df_jornal} conducted field measurements to directly measure in-situ parameters of debris flows at high spatial and temporal resolution with time-lapse LiDARs and cameras. Spielmann \etal~\cite{Spielmann2023debris} manually measure the front velocity and track individual features such as large boulders and woody debris in the point cloud data. Aaron \etal~\cite{jordan2023debris} start from images and explore both pixel-level motion and object-level motion: the former method applies particle image velocimetry (PIV) to the video frames, while the latter focuses on tracking boulders and woody debris using CNN-based object detection. Subsequently, both methods project the 2D displacements into the point clouds. Although both LiDAR and camera sensors are employed, the connection between them is not fully exploited with a simple projection. Further, the velocity fields obtained are relatively sparse when compared to what could be obtained with dense optical flow techniques. To this end, we develop tailored sensor-fusion methods to fully utilize the multi-sensor setup for the debris-flow dataset, resulting in much denser flow velocity fields.

\section{Method}
\label{sec:method}
\begin{figure*}[t!]
     \centering
        \includegraphics[width=1.0\linewidth]{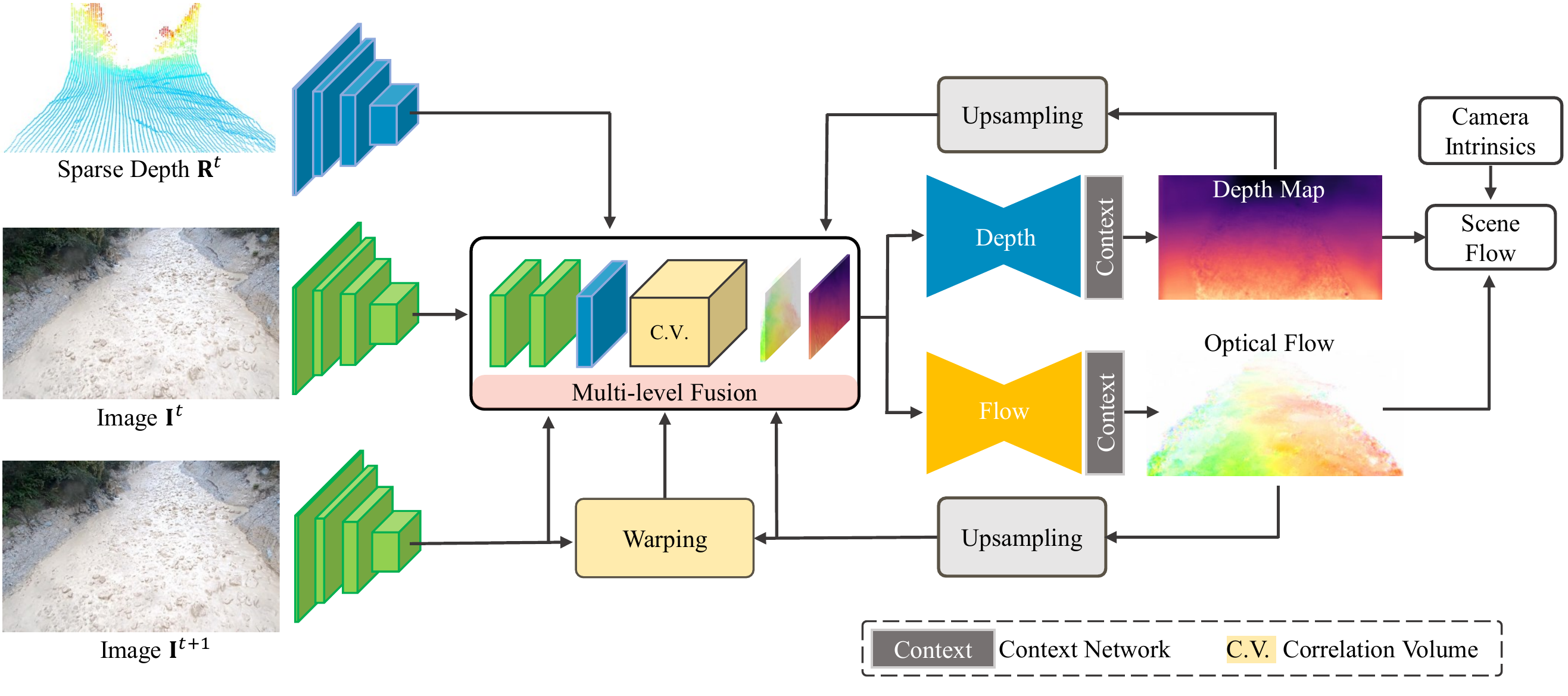}
        \caption{\textbf{Architecture overview}. The inputs of our network are two consecutive camera images and the corresponding (synchronous) range maps generated by a LiDAR sensor. The optical and depth information are encoded by image and sparse depth encoders. Next, a multi-level sensor fusion scheme combines depth features, image features, and the feature correlation volume. The aggregated feature maps are fed into a multi-task decoder to output the depth and optical flow estimates. After the learnable part, we employ deterministic geometric relations to back-project the depth image into the point cloud and compute 3D motion from correspondences defined by the optical flow.}
 	    \vspace{-2mm}
        \label{fig:architecture}
\end{figure*}
The workflow of our method is illustrated in \cref{fig:architecture}. Our network starts with two feature encoders, followed by feature warping, correlation volume computation, and multi-level fusion, and finally decodes the resulting latent features to optical flow and depth estimates.
\cref{sec:optical_flow} and \cref{sec:depth} elaborate two necessary losses to supervise the learning of optical flow and depth without ground truth. In \cref{sec:motion_seg} and \cref{sec:cycle_loss}, we explain how we leverage two inductive biases specific to debris flow. To finally obtain the 3D motion, post-processing (\cref{sec:back_proj}) is used to convert the optical flow and depth to 3D flow. The flow speed profile over time is estimated after multi-frame temporal smoothing (\cref{sec:temp_smooth}). We train the model end-to-end with a loss $\loss$ composed of four terms: 
\begin{equation}
    \loss = \lambda_{\text{flow}}\loss_\flow + \lambda_{\text{depth}}\loss_\depth + \loss_\static + \loss_\cycle\;.
    \label{eq:loss}
\end{equation}
\paragraph{Problem formulation}
Consider an image sequence $\ImageSet = \{\Image^t  \}_{t=1}^T$ and a point cloud sequence $\PCset = \{\PC^t \}_{t=1}^T$ with constant interval in $T$ epochs.
Our objective is to estimate the 3D coordinates $\mathbf{P}^t = (x,y,z)$ and the flow vector $ \mathbf{V}^t = (\Delta x, \Delta y, \Delta z)$ for every pixel $\mathbf{p}^t=(u,v) \in \Image^t$. Additionally, the flow velocity over time needs to be accumulated to generate the flow speed profile $\mathbf{S}=\{\|\mathbf{V}^t\|_2 \}_{t=1}^T$.

\subsection{Backbone network}
\label{sec:backbone_network}
We follow~\cite{hur2020self,sun2018pwc,wu2020pointpwc} and iteratively estimate the optical flow in a coarse-to-fine manner. Our network takes both images and point clouds as inputs, and fuses their features at multiple levels. 

\paragraph{Feature encoders}
The network starts with an image encoder and a depth encoder. Image features $\Feature_{\mathrm{image}}^t$ and  $\Feature_{\mathrm{image}}^{t+1}$ are extracted by a shared image encoder. The depth encoder learns to extract and aggregate the sparse range features $\Feature_{\mathrm{depth}}^t$. Both encoders have a pyramid structure with five layers to preserve global context and local details. The feature pyramid channels are $[32, 64, 96, 128, 192]$ and $[8, 16, 24, 32, 64]$ for image encoder and depth encoder, respectively. At each layer a stride of 2 halves the size of the feature map.

\paragraph{Multi-level fusion}
We carry out feature fusion at each level of the feature pyramid. The first iteration initializes optical flow $\OF^t$ and depth $\mathbf{D}^t$ to \textit{zero}. Based on the current optical flow estimates, the feature map $\Feature_{\mathrm{image}}^{t+1}$ is warped~\cite{wolberg1990warping} back to $t$ to obtain $\tilde{\Feature}_{\mathrm{image}}^{t}$. Then we compute the correlation volume $\Feature_{\mathrm{corr}}^t$ by stacking the feature-wise correlation, between $\Feature_{\mathrm{image}}^t$ and  $\tilde{\Feature}_{\mathrm{image}}^{t}$ with window size $l$. The feature-wise correlation $\feature_{\mathrm{corr}}$ between $\feature^t_i \in \Feature_{\mathrm{image}}^t$ and $\Tilde{\feature}^t_j \in \tilde{\Feature}_{\mathrm{image}}^{t}$ is computed as 
\begin{equation}
\label{eq:sf}
    \feature_{\mathrm{corr}} = \frac{1}{C} \sum_{m \in [0,C)} [\feature^t_i \circ \Tilde{\feature}^t_j]_m\;,
\end{equation}
where $\circ$ denotes the element-wise product, indices $i,j$ refer to the $j^{th}$ correlation of the $i^{th}$ feature, and $C$ is the channel depth of the features. For a feature map $\Feature_{\mathrm{image}}^t \in \mathbb{R}^ {W \times H \times N}$, the correlation volume $\Feature_{\mathrm{corr}}^t \in \mathbb{R}^{W \times H \times l^2}$ is generated. $W$, $H$ and $N$ correspond to the width, height, and number of channels of the feature map, respectively.
Then all the latent features are concatenated,
\begin{equation}
    \label{eq:point-motion}
    \Feature_{\mathrm{fuse}}^t =\cat(\Feature_{\mathrm{image}}^{t},\Feature_{\mathrm{image}}^{t+1},\Feature_{\mathrm{depth}}^{t},\Feature_{\mathrm{corr}}^t,\mathbf{O}^t,\mathbf{D}^t)\;,
\end{equation}
and the fused feature map $\Feature_{\mathrm{fuse}}^t$ is fed into the decoders. 

\paragraph{Decoder design}
We adopt separate decoders for depth and flow as described in ~\cite{hur2021self}, for better convergence. The fused feature $\Feature_{\mathrm{fuse}}^t$ first goes through a shared convolutional layer to output a latent feature map, which is then passed into two separate decoders. The context network from~\cite{yu2015multi,chen2017deeplab} is adopted after each decoder, which outputs the depth and flow estimates.

\subsection{Optical flow estimation}
\label{sec:optical_flow}
The optical flow decoder consists of three convolutional layers and a context network (dilated convolutions). Unlike vehicles in KITTI~\cite{geiger2012we} or NuScenes~\cite{caesar2020nuscenes}, debris-flow motion is non-rigid. It is hard to model debris-flow motion through the Navier-Stokes equation~\cite{constantin2020navier} due to significant unknowns related to the underlying mechanisms that control flow properties. Moreover, manual annotation of optical flow on such a dataset is almost infeasible. Therefore, we use an unsupervised loss~\cite{jonschkowski2020matters,baker2011database} based on photometric consistency and local smoothness. Given two images $(\Image^t, \Image^{t+1})$ and an optical flow $\OF^t_{f}$, we warp $\Image^{t+1}$ back to $t$: 
\begin{equation}
   \Tilde{\Image}^{t} = \omega(\Image^{t+1}, \OF^t_{f})\;,
\end{equation}
where $\omega(\cdot,\cdot)$ represents the backward warping function with the first argument being the scalar field to be warped and the second argument the warping correspondences. Under the assumption that the appearance of the same object remains constant, the photometric loss $\loss_\photo$ penalizes the photometric differences between $\Image^{t}$ and $\Tilde{\Image}^{t}$:
\begin{equation}
    \loss_\photo = \operatorname{SSIM}\left(\mathrm{\phi}(\Image^{t}),\mathrm{\phi}(\Tilde{\Image}^{t})\right),
\end{equation}
where $\phi(\cdot)$ is an average pooling module and  $\operatorname{SSIM}(\cdot,\cdot)$ is the structural similarity index~\cite{wang2004ssim}.

Besides, a first-order regularizer~\cite{baker2011database} $\loss_\smooth$ encourages edge-aware smoothness~\cite{woodford2009global,meister2018unflow,hur2020self} of the estimates by penalizing the sum of first-order derivatives, weighted by the image gradients:
\begin{equation}
\resizebox{.95\hsize}{!}{$\loss_\smooth = \frac{1}{H \times W \times C} \sum_{\mathbf{p}} \sum_{i \in \mathbf{I}^t}\left\|\nabla_i^1 \OF^t(\mathbf{p})\right\|_1 \cdot e^{-\beta\left\|\nabla_i \mathbf{I}^t(\mathbf{p})\right\|_1}$},
    \label{eq:smoothness}
\end{equation}
 with $\beta = 10$ and $H \times W \times C$ the image dimensions.
The total flow loss is $\loss_\flow = \loss_\photo + \lambda\loss_\smooth$, with $\lambda = 0.15$. In the forward propagation, we additionally swap the two input images~\cite{sun2018pwc,hur2020self} and compute the backward flow $\OF^t_b$ and its corresponding photometric loss. During training $\loss_\flow$ includes the bidirectional flow loss, whereas only forward flow is computed at inference time. In this way we double the training samples, moreover the bidirectional flows are vital to leverage fluid continuity (\cref{sec:cycle_loss}).

\subsection{Depth estimation}
\label{sec:depth}
\begin{figure}[h]
     \centering
        \includegraphics[width=1.0\linewidth]{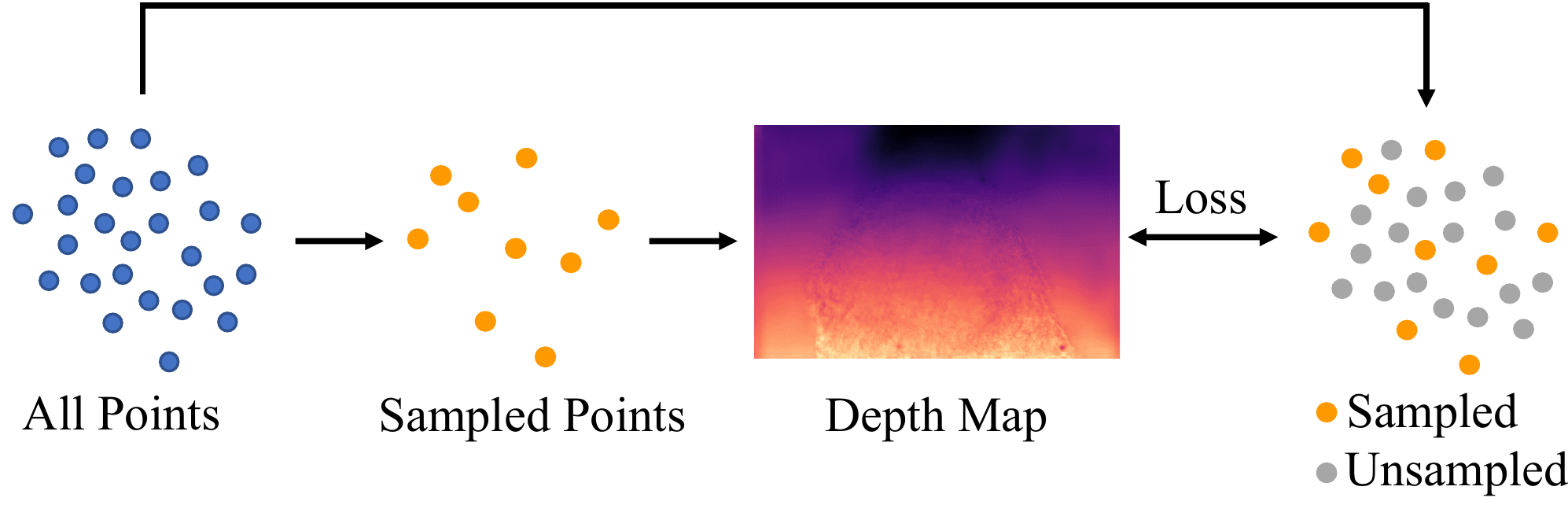}
       \caption{\textbf{A simple and effective way of supervising depth}. A downsampled point cloud serves as input, the full point cloud as the target.}
   \label{fig:lidar_sup}
\end{figure}
Previous work on unsupervised and zero-shot monocular depth estimation~\cite{godard2017monodepth1,godard2019monodepth2,ranftl2020zeroshotmono,hur2020self} and depth completion~\cite{Wong2021completion} does not rely on ground truth (direct depth) as supervision, but instead uses disparity maps acquired from stereo images. In the debris flow dataset, we have neither ground truth nor good stereo pairs. Therefore, we propose a novel way to supervise the depth estimation as shown in \cref{fig:lidar_sup}. We start from randomly downsampling the input point cloud $\PC^t$ into $\PC^t_{down}=[\point_i^t,...,\point_j^t,...,\point_{n_t}^t] \in \mathbb{R}^{3 \times (\eta \times n_t)}$ with $\eta$ being the downsampling ratio. During training, only the sampled point cloud $\PC^t_{down}$ is converted to a range map $\mathbf{R}^t_{down}$, and the network makes predictions based on that downsampled map. During back-propagation, the original input point cloud $\PC^t$ serves as ground truth. After converting $\PC^t$ into a sparse range map $\mathbf{R}^t$, we bilinearly sample the points with a valid LiDAR footprint from the dense prediction $\mathbf{D}^t$ and compute the loss from those samples and the sparse range map:
\begin{equation}
\resizebox{1.0\hsize}{!}{$\loss_1 = \frac{1}{H \times W \times C} \sum_{i} \left\|\mathbf{R}^t(i) -\xi\big( \mathbf{\Theta}_{depth}(\Image^t, \Image^{t+1},\mathbf{R}^t_{down}), \mathbf{R}^t(i) \big)\right\|_1$},
    \label{eq:depth_l1} 
\end{equation}
where $\mathbf{\Theta}_{depth}$ represents the depth branch of our network and $\mathbf{R}^t_{down}$ is the downsampled range map. The bilinear resampling function $\xi(\mathbf{X},y)$ retrieves $x$ from $\mathbf{X}$ at the position of $y$. As in \cref{sec:optical_flow}, smoothness regularization $\loss_\smooth$ is used in the depth prediction. The depth estimation branch is supervised by the weighted sum 
$\loss_\depth = \loss_1 + \lambda \loss_\smooth$, with $\lambda = 0.1$. The rationale behind this supervision strategy is that \textit{(i)} the sampled LiDAR points teach the network to preserve the ground truth input values, while \textit{(ii)} remaining LiDAR points provide a supervision signal for the densification/interpolation from sparse range input to dense output.

\subsection{Motion segmentation}
\label{sec:motion_seg}
The debris flow dataset is captured by a static monitoring system. Different from many scene flow methods~\cite{gojcic2021weakly,huang2022dynamic} which subtract the ego-motion before analyzing the dynamic scene, we leverage the prior of \textit{zero} ego-motion, via a loss $\loss_\static$. As described in \cref{sec:optical_flow}, we compute the forward flow
\begin{equation}
    (\OF^t_f, \mathbf{D}^t) = \mathbf{\Theta}(\Image^t, \Image^{t+1},\mathbf{R}^t_{down})\;,
\end{equation}
and the backward flow by swapping the input order 
\begin{equation}
    (\OF^t_b, \mathbf{D}^{t+1}) = \mathbf{\Theta}(\Image^{t+1}, \Image^t,\mathbf{R}^{t+1}_{down})\;,
\end{equation}
where $\mathbf{\Theta}$ are the parameters of the network. With bidirectional flows $\OF^t_f$ and $\OF^t_b$ and depth estimates $\mathbf{D}^t$ and $\mathbf{D}^{t+1}$, we define a prior that favours a static sensor system. After distinguishing static pixels from moving ones with a simple threshold $\varepsilon$, the loss demands that depth estimates in the static part should be consistent, and penalizes deviations from that soft constraint:
\begin{equation}
    \loss_\static = \frac{\sum_{\mathbf{p}} \left(\Mask^{t}(\mathbf{p}) \cdot \left\|\mathbf{D}^{t+1}(\mathbf{p}) -  \mathbf{D}^{t}(\mathbf{p})\right\|_1 \right)}
    {\sum_{\mathbf{q}} \Mask^{t}(\mathbf{q})}\;.
\end{equation}
Here $\mathbf{p,q}$ represents pixel coordinates in the image, and $\Mask^{t}$ denotes a binary mask that is 1 for static pixels and 0 for moving ones. The loss $\loss_\static$ builds a bridge so that optical flow estimation and depth estimation can mutually supervise each other. Inaccurate optical flow estimates lead to wrong motion segmentation, and consequently, $\loss_\static$ increases due to the misalignment of non-static pixels. Unstable or inaccurate depth estimates in static regions also cause the increase of $\loss_\static$.

\subsection{Fluid continuity}
\label{sec:cycle_loss}
One long-standing problem in optical flow estimation is occlusion. 
In this work, we use the inconsistency~\cite{meister2018unflow,xu2022gmflow,xu2022unifying} of forward-backward optical flow to achieve an inductive bias towards fluid continuity. The sensor system is static, and the motion only occurs on the fluid surface with approximately in-plane motion. The continuity of the fluid surface results in almost no occlusion, so we integrate this prior into the loss $\loss_\cycle$ as
\begin{equation}
    \loss_\cycle = \frac{1}{H \times W \times C} \sum_{\mathbf{p}} \left\| \OF_f^t(\mathbf{p}) + \omega(\OF_b^t,\OF_f^t)(\mathbf{p})   \right\|_1 \;,
\end{equation}
with $\omega(\cdot,\cdot)$ the backward warping function from \cref{sec:optical_flow}.

\subsection{Multi-frame temporal smoothing}
\label{sec:temp_smooth}
\begin{figure}[t]
     \centering
        \includegraphics[width=0.7\linewidth]{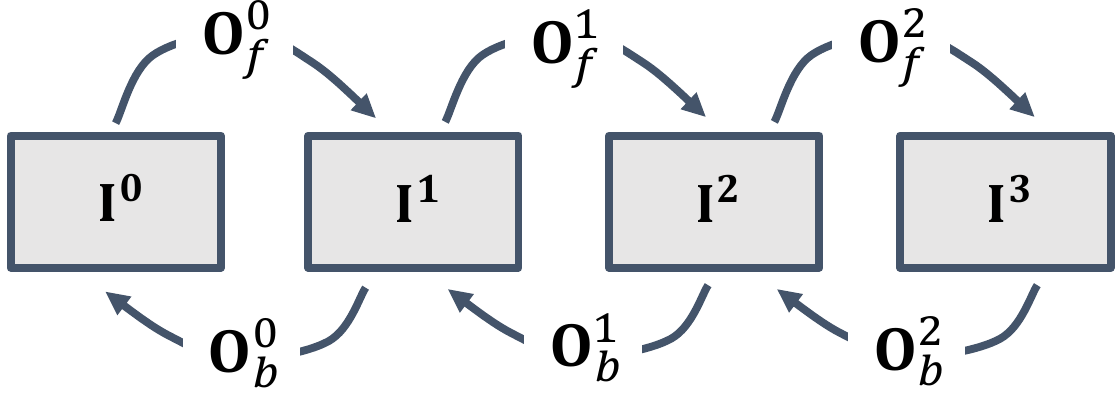}
   \vspace{-2mm}
       \caption{\textbf{Schematic illustration of  temporal smoothing for flow speed estimation.}}
    \vspace{-4mm}
   \label{fig:temporal_smoothing}
\end{figure}

Debris flow motion analysis, \eg, calculating flow speed profiles over time, requires more than two-frames. We add a post-processing module to achieve a temporally smoother flow speed estimate, as shown in \cref{fig:temporal_smoothing}. 
Given consecutive images $\{ \mathbf{I}^t\}_{t=0}^3$, and their bidirectional flow estimates $\{\OF^m_n\}_{n=f,b}^{m=0,1,2}$, our goal is to smooth the flow $\OF^1_f$. The assumption is that the velocity of object points remains constant over a short time interval. We start with warping $\OF_f^0$ forward to the next frame $ \OF^{1'}_f = \omega(\OF^{0}_f, \OF^{0}_b )$. Then $\OF_f^2$ is warped backward $\OF^{1''}_f = \omega(\OF^{2}_f, \OF^{1}_f )$. We compute the smoothed $\Tilde{\OF}^{1}_f$ as
\begin{equation}
    \Tilde{\OF}^{1}_f = \lambda_0 \OF^{1'}_f + \lambda_1 \OF^{1}_f + \lambda_2 \OF^{1''}_f\;,
\end{equation}
where $\lambda_0+\lambda_1+\lambda_2=1$. In the end, we apply temporal smoothing to the per-frame estimates to generate the flow speed profile.

\subsection{Lifting optical flow to 3D}
\label{sec:back_proj}
As a final step we transform optical flow to 3D scene flow with the help of depth. The camera matrix $\K_{3\times3}$ and the camera-LiDAR transformation matrix $\T_{3\times4}:[\mathbf{R} \mid \mathbf{t}]$ are known from calibration. Given a depth estimate $z_i^t$, every pixel $\mathbf{p}^t_i \in \mathbf{I}^t$ is back-projected into a 3D point $ \Point^t_i$ as
\begin{equation}
\Point_i^t  =   z_i^t \mathbf{R}^{\mathrm{T}} \mathbf{K}^{-1} \mathbf{p}^t_i - \mathbf{R}^{\mathrm{T}} \mathbf{t}\;.
\label{eq:projection}
\end{equation}
The same transformation is applied to the second frame $\mathbf{I}^{t+1}$. Consider two point clouds $\{\Point^t_i\}_{i=0}^{\mathrm{w\times h}}$ and $\{\Point^{t+1}_i\}_{i=0}^{\mathrm{w\times h}}$, we determine the motion for every point in frame $t$ by subtracting its positions in epoch $t+1$ and $t$ as in \cref{eq:sf}.

\subsection{Implementation details}
We have implemented our network in PyTorch~\cite{paszke2019pytorch} and trained it from scratch on a single RTX 6000/3090 24G GPU. During training the loss of \cref{eq:loss} is minimized with the Adam optimizer~\cite{kingma2014adam}, with hyper-parameter $\beta_1 = 0.9$ and an initial learning rate of 0.0004. We train the network for 30 epochs with a batch size of 8, the learning rate decays by half after 5, 7 and 15 epochs. To enlarge the training set, we adopt geometric data augmentation, including random cropping and flipping for images and range maps simultaneously. 
The original image size is $1920 \times 1080$, which we crop it to $1600 \times 960$, making sure the dimensions are divisible by $2^5=32$.

\section{Experiments}
\label{sec:results}
We first introduce the debris flow dataset and our evaluation metrics for optical flow and depth estimation. Then we compare our model to several baselines. Finally, we perform an ablation study to validate our network architecture and loss design. 

\subsection{Debris flow dataset}
\label{sec:debris_dataset}
The debris flow event was captured by a camera-LiDAR setup at Illgraben, Switzerland~\cite{mcardell2021illgraben}. The whole event lasted for approximately 30 minutes and consisted of three stages. Stage \textit{(i)} features a watery pre-surge, stage \textit{(ii)} marks the arrival of the flow front with boulders and woody debris and is followed by stage \textit{(iii)} with a finer-grained slurry. This event is described and analysed in detail by ~\cite{jordan2023debris,Spielmann2023debris,aaron2023df_jornal}. The point clouds are captured by a 64-beam Ouster OS1-64 (Gen-1) LiDAR at a frequency of 10 Hz with each scanline consisting of 2048 points. Along with the LiDAR, two cameras are also deployed, capturing high-resolution videos of the event at 25 Hz. The rigid body transformation between the two sensors as well as the camera intrinsics are known from calibration. The dataset consists of a total of 6000 ordered frames (\ie, the first 10 minutes of the flow). We divide it into chunks of 60 frames, of which the first 45 are assigned to the training set and the last 15 are used for evaluation. This split avoids excessive (temporal) correlation between the training and test sets, but still allows one to evaluate model performance at different stages of the debris flow event. As the two cameras are poorly synchronized and do not form a good stereo baseline, we build and evaluate our method only with the top-view camera and LiDAR.

\subsection{Evaluation setting}
\label{sec:dataset_evalsetting}

\paragraph{Optical flow metrics}
As our dataset lacks a direct ground truth for optical flow, typical metrics such as angular error (AE)~\cite{baker2011database} or end-point error (EPE)~\cite{nagel1986investigation} are not applicable; while $\loss_\photo$ is not an independent metric, since it is part of the loss minimized during training. We resort to the frame interpolation metric of Baker \etal~\cite{baker2011database}.
With the optical flow estimate $\mathbf{O}^t_f$, we reconstruct the first frame $\Tilde{\Image}^{t}$ by back-warping the second frame $\Image^{t+1}$. The root-mean-square deviation (RMSD) between the original image and the reconstructed image is computed as 
\begin{equation}
    \mathrm{RMSD}=\sqrt{\frac{1}{H \times W \times C} \sum_\mathbf{p}\left|\Image^{t}(\mathbf{p})-\Tilde{\Image}^{t}(\mathbf{p})\right|_2^2}\;.
\end{equation}
Furthermore, we also utilize the ternary census transform loss~\cite{meister2018unflow,zou2018df,stein2004efficient} in our experimental analysis, as it is robust against changes in illumination.

\paragraph{Depth estimation metrics} For the evaluation of depth estimates, we use the depths from LiDAR points as ground truth and compute the mean absolute error within 10 m, 30 m, and 50 m distance from the LiDAR sensor ($\mathrm{MAE_{10}}$, $\mathrm{MAE_{30}}$ and $\mathrm{MAE_{50}}$) as well as the absolute relative error ($\mathrm{Abs.\: Rel.}$).

\paragraph{Baselines}
We establish four baselines for comparison purposes. For optical flow estimation, we employ RAFT~\cite{teed2020raft}, a leading supervised method, and compare its pretrained and finetuned model to ours. We also include a comparison with an unsupervised optical flow network~\cite{meister2018unflow} as another baseline. To evaluate depth estimation, we construct an image-only baseline, by removing the depth encoder branch in our network. We also use IP-Basic~\cite{ku2018defense} as another depth completion baseline. Lastly, we establish the final baseline by comparing our estimates to the PIV-based flow speed estimates of Aaron~\etal~\cite{jordan2023debris} for the same dataset.

\subsection{Main results}
\label{sec:evaluation_results}

\begin{table}[h]
\centering
\setlength{\tabcolsep}{10pt}
\resizebox{\columnwidth}{!}{
 \begin{tabular}{lcc}
\toprule \multicolumn{1}{l}{Method} & RMSD $\downarrow$ & Census Loss $\downarrow$\\
\midrule 
\multicolumn{1}{l}{UnFlow~\cite{meister2018unflow}}  &  19.58 & 0.146\\
\multicolumn{1}{l}{RAFT-Sintel~\cite{teed2020raft}} & 11.75 & 0.156 \\
\multicolumn{1}{l}{RAFT-Sintel-ft~\cite{teed2020raft}}  &  10.18 & 0.147\\
\multicolumn{1}{l}{DeFlow-Cam (\textit{Ours})}  & \underline{7.25} & \underline{0.109} \\
\multicolumn{1}{l}{DeFlow-Fusion (\textit{Ours})}  &\textbf{7.02} & \textbf{0.106} \\

\bottomrule
\end{tabular}
}
\vspace{-2mm}
\caption{\textbf{Results of optical flow estimation.}}
\vspace{-4mm}
\label{tab:of_baseline}
\end{table}

\paragraph{Optical flow estimation} We test unsupervised UnFlow~\cite{meister2018unflow}, which shows weak performance on the debris flow dataset. We also evaluate a state-of-the-art optical flow method, RAFT~\cite{teed2020raft}, on the debris flow dataset. We first evaluate the model pretrained on Sintel~\cite{Butler_sintel}. 
Furthermore, we also finetuned that model, using the same flow loss $\loss_\flow$ and the same hyperparameter settings as in our new method, to conduct a fair comparison. After finetuning, we can see a performance gain of RAFT on the debris flow dataset (\cref{tab:of_baseline}), which confirms a domain gap between canonical flow datasets and the newly available debris flow. 
Both our hard baseline DeFlow-Cam and full model DeFlow-Fusion outperform the finetuned RAFT model by a large margin of $\approx 30\%$, see \cref{tab:of_baseline}. In terms of network design, our architecture internally downsamples the raw input by a factor $\times$2, while RAFT downsamples to $\times$8 the original size. Both RAFT and our network adopt an iterative refinement scheme. We believe that the salient and variable local texture information in debris flow is better preserved by our model due to less downsampling and consequently higher-resolution feature maps.

\paragraph{Depth estimation}
We initiated our study with a camera-only baseline similar to \cite{hur2020self} without the top depth encoder in \cref{fig:architecture} to assess whether a single camera is capable of providing sufficient information for the task. In \cref{tab:mono_depth}, we denote the camera-only model and the camera-LiDAR fusion as \textit{DeFlow-Cam} and \textit{DeFlow-Fusion}, respectively. To ensure a fair comparison, we trained both models using only the two essential loss terms $\loss_\flow$ and $\loss_\depth$.
The DeFlow-Fusion model achieves sub-decimeter accuracy, surpassing the performance of DeFlow-Cam significantly. By fusing the depth branch into our network, the $\mathrm{Abs.\: Rel}$ is reduced 6$\times$, from 6.3\% to 0.9\%. 
DeFlow-Fusion also outperforms the depth completion baseline IP-Basic~\cite{ku2018defense} by approximately 30\% in all metrics.
Thus, we conclude that a camera-LiDAR fusion approach is essential for debris flow estimation, where the sparse LiDAR input provides strong guidance for depth estimation. As shown in \cref{tab:of_baseline}, the integration of LiDAR not only improves depth estimation, but also enhances the optical flow. We believe this performance gain is due to the LiDAR branch and sensor fusion module providing more information about the 3D scene structure, which allows the network to better understand debris-flow motion. 
\begin{table}[h]
\centering
\renewcommand{\arraystretch}{1.3}
\setlength{\tabcolsep}{2pt}
\resizebox{1.0\columnwidth}{!}{
 \begin{tabular}{lcccc}
\toprule 
 Method  & $\mathrm{MAE_{10}}[m]\downarrow$ & $\mathrm{MAE_{30}}[m]\downarrow$ & $\mathrm{MAE_{50}}[m]\downarrow$ & $\mathrm{Abs.\: Rel. [\%]}\downarrow$\\
\midrule 
IP-Basic~\cite{ku2018defense}  & 0.06 & 0.17 & 0.22 & 1.1 \\
DeFlow-Cam  & 0.38  & 0.83  & 1.02  & 6.3 \\
DeFlow-Fusion  & \textbf{0.04}  & \textbf{0.10}  & \textbf{0.14}  & \textbf{0.6} \\

\bottomrule
\end{tabular}
}
\vspace{-1mm}
\caption{ \textbf{Depth estimation on our debris flow dataset.}}
\vspace{-4mm}
\label{tab:mono_depth}
\end{table}

\begin{figure}[t]
     \centering
        \includegraphics[width=0.96\linewidth]{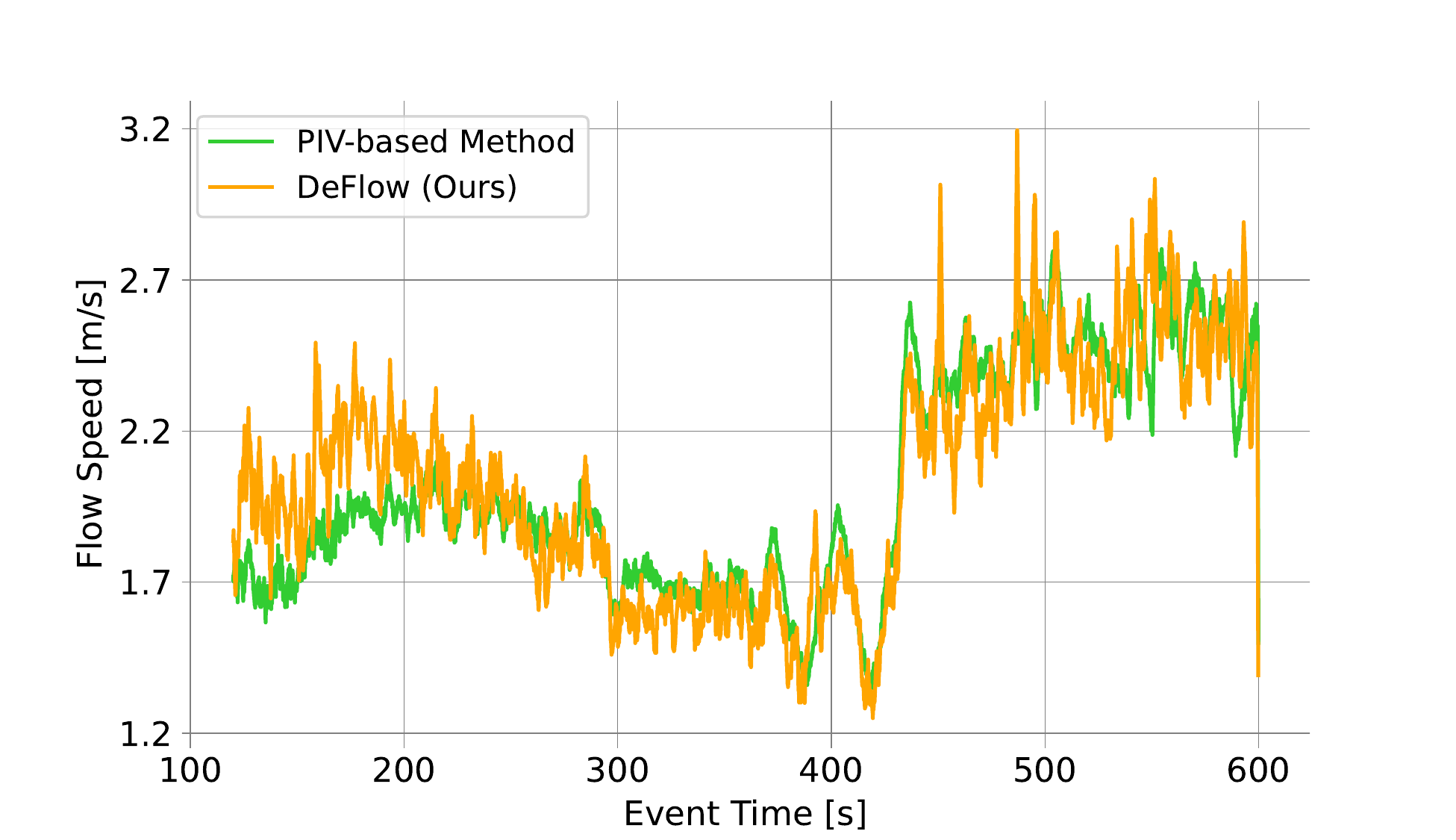}
        \caption{\textbf{Flow speed estimates over time.} Comparison between our method (DeFlow) and PIV method of Aaron \etal~\cite{jordan2023debris,aaron2023df_jornal}.}
   \label{fig:flow_speed}
   \vspace{-0.2cm}
\end{figure}

\paragraph{Flow speed estimation}
Our approach determines the pixel-wise 3D motion for every frame with multi-frame temporal smoothing. It can therefore provide velocity and speed estimates for any region in the flow channel. Here we compute an estimate of the flow speed within a vertical bounding box $\{(x,y,z)\mid x\in[-1.0, 1.0],y \in[19.0,21.0]\}$ in the LiDAR frame, the same region also used in~\cite{jordan2023debris}. The average flow speed in the bounding box during the whole event is computed and compared to the results of Aaron \etal~\cite{jordan2023debris}, derived with a PIV-based method and validated with manual feature measurements (\cref{fig:flow_speed}). The two results are similar in terms of the overall trend, the absolute velocities, as well as the timing of peaks.
\begin{figure}[h]
     \centering
        \includegraphics[width=1\linewidth]{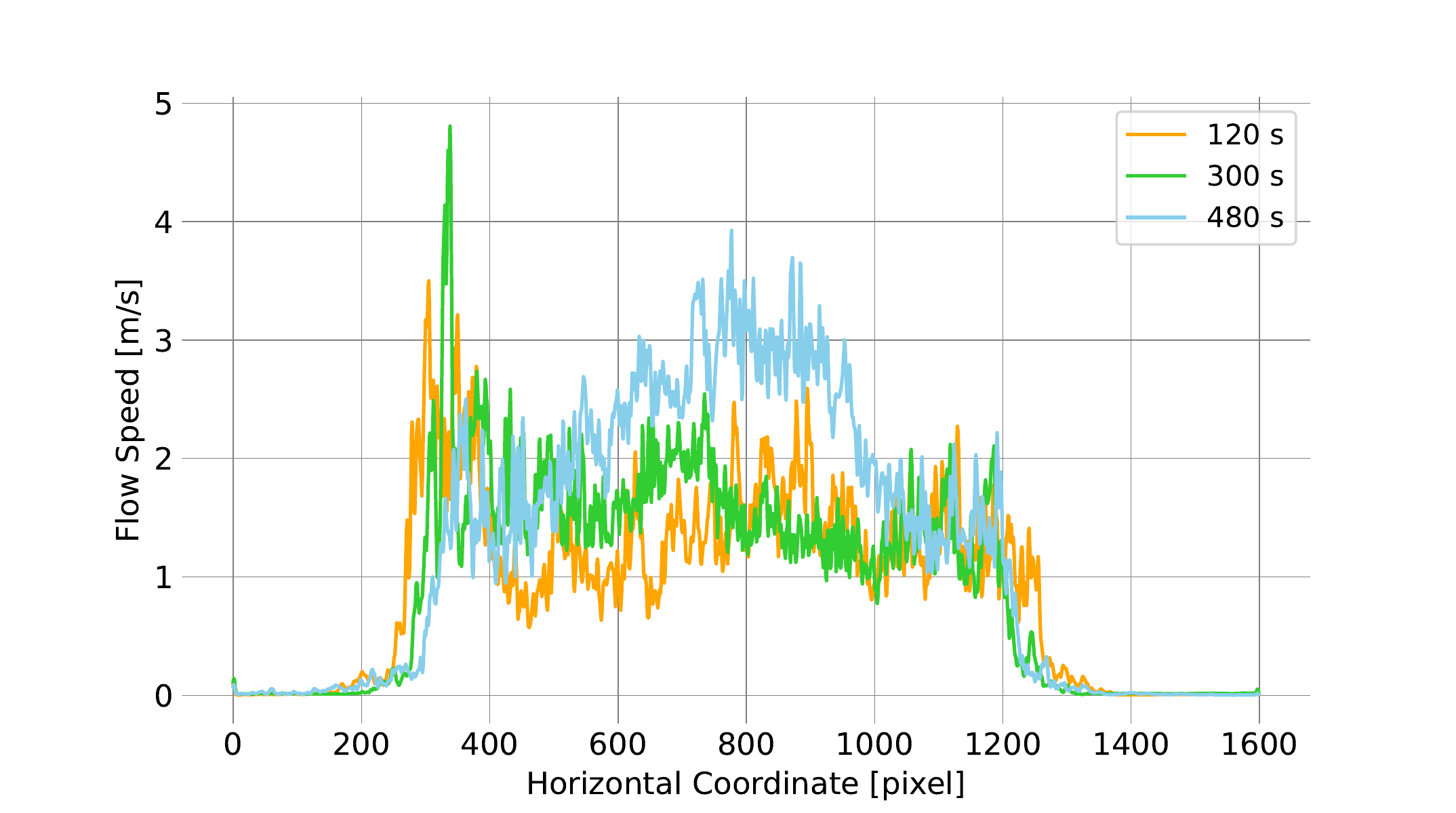}
       \caption{\textbf{Spatial distribution of flow speeds across the channel.}}
       \vspace{-4mm}
   \label{fig:channel_profile}
\end{figure}

\paragraph{Channel speed distribution}
To better understand how flow velocity is distributed in space, we investigate the flow speed along a horizontal cross-section of the channel. Specifically, we plot the velocity against the x-axis with $y \in [470,490]$ at three different times, represented by epoch 1200, 3000, and 4800 (\cref{fig:channel_profile}). During the first two epochs (1200 and 3000), we observe a nearly constant cross-channel velocity profile, whereas during the final epoch the shape of the velocity profile changes.  This corresponds to a large rise in overall flow velocity (\cref{fig:flow_speed}), and may reflect changes in flow composition~\cite{aaron2023df_jornal}.

\paragraph{Qualitative results}
\begin{figure*}[t]
    \centering
    \includegraphics[width=1.0\linewidth]{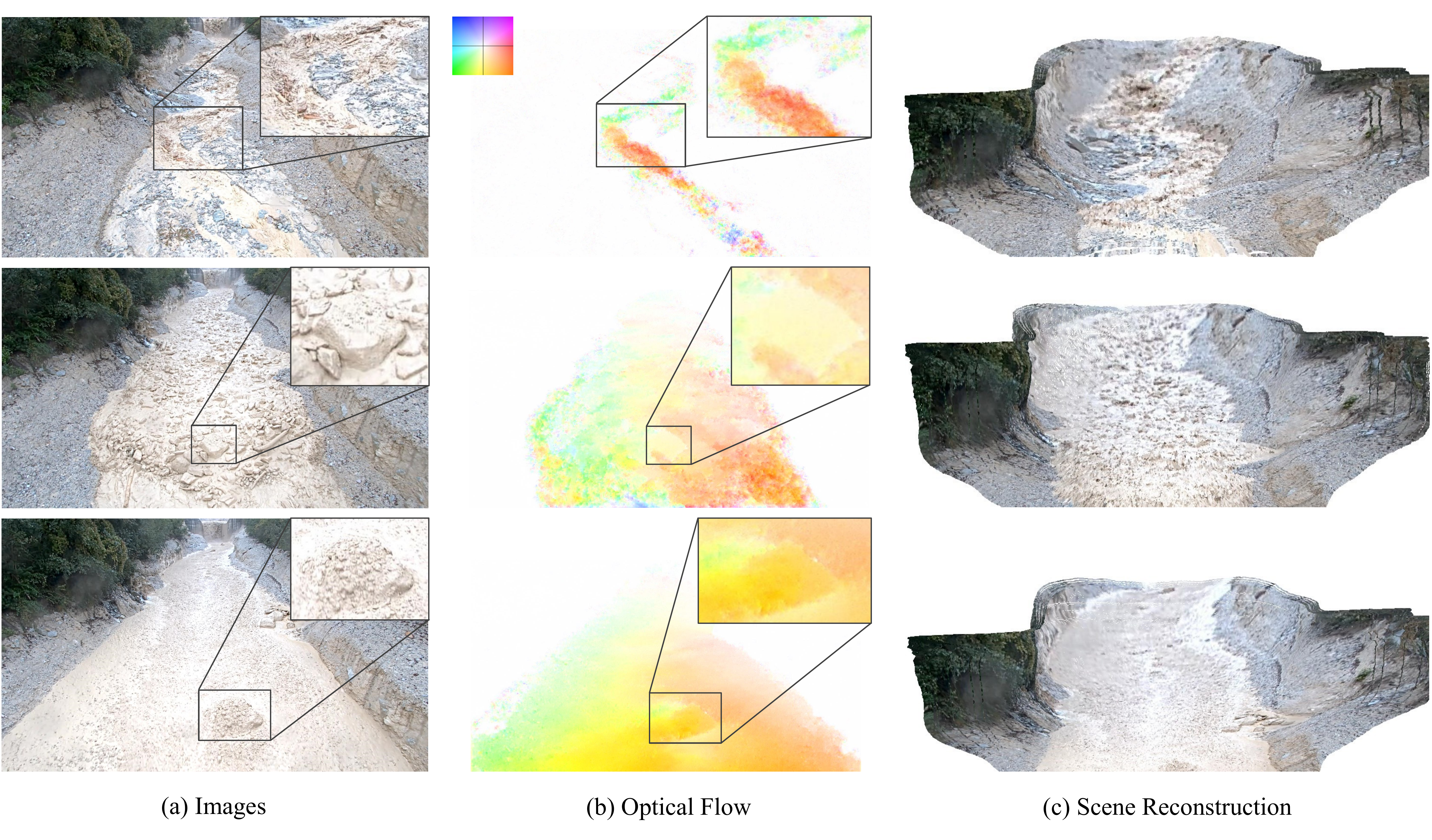}
    \vspace{-4mm}
    \caption{\textbf{Qualitative results}. Raw images (a), optical flow estimates (b), and scene reconstruction (c).}
    \vspace{-2mm}
    \label{fig:qualitative_results}
\end{figure*}
We show qualitative examples of optical flow fields and scene reconstructions generated by our approach. Our model successfully segments the motion of large features, such as boulders, in the flow field (\cref{fig:qualitative_results}).  It further distinguishes between dynamic and stationary regions, and reconstructs the 3D scene geometry, as shown in \cref{fig:qualitative_results}. These results represent the densest flow field obtained so far for a natural debris flow, and further processing will make it possible to better understand the interaction of large particles with the surrounding fine-grained slurry.

\subsection{Ablation Study}
\label{sec:ablation}
\begin{table}[h]
    \setlength{\tabcolsep}{2pt}
    \renewcommand{\arraystretch}{1.4}
    \centering	
    \resizebox{1.0\columnwidth}{!}{
    \begin{tabular}{cccc|c|ccccc}
    \toprule
      & \multicolumn{3}{c|}{Modules} & \multicolumn{1}{c|}{Optical Flow$\downarrow$} &   \multicolumn{4}{c}{Depth Estimation$\downarrow$} \\
    & Context & $\loss_\static$   & $\loss_\cycle$ & RMSD & $\mathrm{MAE_{10}}[m]$ & $\mathrm{MAE_{30}[m]}$ & $\mathrm{MAE_{50}[m]}$  & $\mathrm{Abs.\: Rel. [\%]}$ \\
    \midrule
    & \ding{56} & \ding{56}& \ding{56} &7.018 & 0.053 & 0.139 & 0.196 & 0.83  &\\
    & \ding{52} & \ding{56}& \ding{56} &6.969 & \underline{0.047 }& 0.155 & 0.207 & 0.90  &\\
    & \ding{52} & \ding{56}& \ding{52} &7.024 & 0.051 & 0.132 & \underline{0.179 }&  0.82  &\\
    & \ding{52} & \ding{52}& \ding{56} &\underline{6.871} & 0.049 & 0.137 & 0.196 &  \underline{0.81} &\\
    & \ding{56} & \ding{52}& \ding{52} &6.954 & 0.049 & \underline{0.136 }& 0.186 &  0.82  &\\
    & \ding{52} & \ding{52}& \ding{52} &\textbf{6.833} & \textbf{0.046 }& \textbf{0.130 }& \textbf{0.176 }&  \textbf{0.79 } &\\
  
    \bottomrule
    \end{tabular}
    } 
    \vspace{-1mm}
    \caption{\textbf{Ablation study on network and loss design.} The combination of $\loss_\static$ and $\loss_\cycle$ offers the best performance.} 
    \vspace{-2mm}
    \label{tab:loss_design}
\end{table}
\begin{table}[h]
    \setlength{\tabcolsep}{2pt}
    \renewcommand{\arraystretch}{1.4}
    \centering	
    \resizebox{\columnwidth}{!}{
    \begin{tabular}{c|c|ccccc}
    \toprule
     \multirow{2}{*}{Ratio} &  \multicolumn{1}{c|}{Optical Flow$\downarrow$} &   \multicolumn{4}{c}{Depth Estimation$\downarrow$} \\
    &  RMSD & $\mathrm{MAE_{10}}[m]$ & $\mathrm{MAE_{30}}[m]$ & $\mathrm{MAE_{50}}[m]$  & $\mathrm{Abs.\: Rel.[\%]}$\\
    \midrule
    0.2 & \underline{6.990} & 0.057 & 0.196 & 0.269 & 0.11  &\\
    0.5 & 7.018 & \underline{0.053 }& \underline{0.139 } & \underline{0.196 }& \underline{0.83 }  &\\
    0.8 & \textbf{6.982} & \textbf{0.048 }& \textbf{0.131 }& \textbf{0.183 }& \textbf{0.79 } &\\
    \bottomrule
    \end{tabular}
    } 
    \vspace{-1mm}
    \caption{\textbf{Ablation study on downsampling ratios.}} 
    \vspace{-4mm}
    \label{tab:depth_sampling}
\end{table}
\paragraph{Loss design}
We ablate the two additional loss terms $\loss_\static$ and $\loss_\cycle$ with the context network. The results are presented in \cref{tab:loss_design}, which demonstrate that the model trained with all loss terms performs the best, indicating that each of the losses contributes to overall model performance. We also observe that while the context network (\cref{fig:architecture}) on its own is not particularly helpful, it significantly enhances the model performance when used in conjunction with our loss design.

\paragraph{Depth downsampling}
To validate our downsampling strategy for depth supervision, we train three identical models using different depth input ratios and examine the resulting performance changes, as shown in \cref{tab:depth_sampling}. The results indicate that using a higher input ratio leads to a more precise depth map, suggesting that the network is able to leverage varying levels of depth information and adjust its performance accordingly.

\section{Conclusion}
\label{sec: conclusions}
In this study, we have focused on motion estimation under the unique properties exhibited by debris flows, and have presented a multi-task learning approach that integrates images and point clouds. Our self-supervised method is able to estimate the scene structure (depth) and the motion field at a level of detail that is unprecedented in the context of debris-flow research. Further analysis of these results will provide insights into the mechanisms that govern debris-flow motion, which can ultimately be used to reduce the risk associated with these destructive events. We release the dataset to the vision community, providing a new analysis domain and performance benchmark. For \textit{future work}, one promising avenue is to incorporate physical constraints from fluid and soil mechanics to boost the model training. Another important direction is to develop automated algorithms for infrared cameras, in order to enable monitoring during nighttime.

\paragraph{Acknowledgements} Funding for this project was provided in part by an Swiss National Science Foundation grant to J. Aaron (p 193081).  The authors further thank Stefan Boss, Christoph Graf, Brian McArdell and Raffaele Spielmann for interesting discussions, as well as support in setting up and running the video camera and LiDAR systems.

{\small
\bibliographystyle{ieee_fullname}
\bibliography{egbib.bib}
}

\newpage
\setcounter{section}{0}
\renewcommand\thesection{\Alph{section}}
\clearpage
\section{Appendix}
\beginsupplement
\subsection{Dataset}
\begin{figure}[h!]
     \centering
        \includegraphics[width=1.0\linewidth]{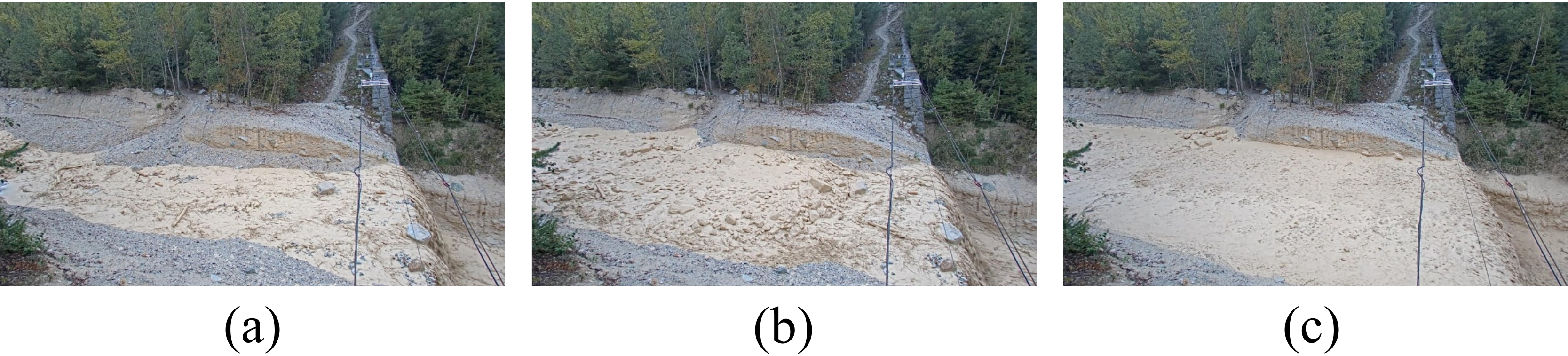}
   \vspace{-4mm}
       \caption{Three stages of debris flow: (a) pre-event, (b) arrival of boulder front and (c) fine-grained slurry fluid.}
   \label{fig:flow_stage}
\end{figure}

\begin{figure}[h!]
     \centering
        \includegraphics[width=1\linewidth]{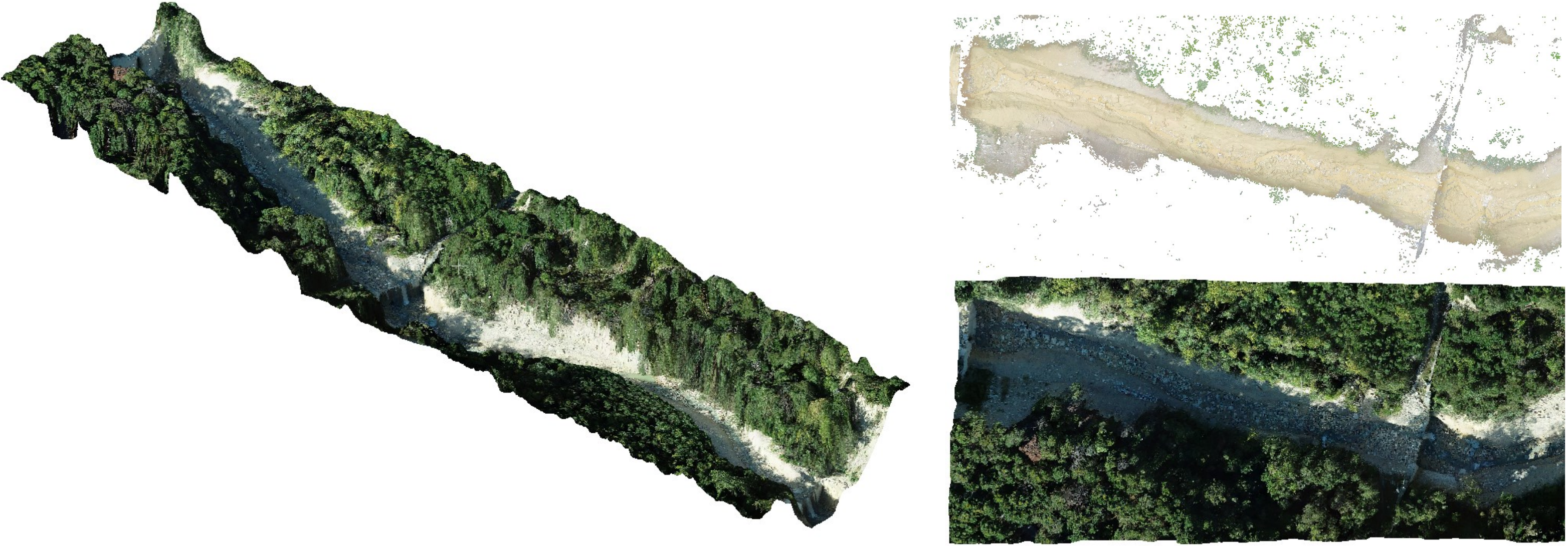}
   \vspace{-2mm}
       \caption{\textbf{3D reconstruction of the debris-flow monitoring site~\cite{de2022flow}.} Overview of the scene on the left. Reconstruction of the channel before the event on the upper right, and after the event on the bottom right. }
    \vspace{-4mm}
   \label{fig:uav}
\end{figure}

In this section, we provide supplementary information and visualizations of the debris flow dataset. The representative shapes of the debris-flow surface are summarized in \cref{fig:flow_stage}. The terrain of the monitoring site is shown in \cref{fig:uav}.

\noindent In the debris flow dataset, we release:
\begin{itemize}[noitemsep] 
    \item 6000 high-resolution images (1920$\times$1080) 
    \item 6000 high-accuracy point clouds
    \item Camera calibration matrices
    \item Rotation matrices between camera and LiDAR
    \item Translation vectors between camera and LiDAR
\end{itemize}

\subsection{Evaluation Metrics}
\paragraph{SSIM} Structural similarity index~\cite{wang2004ssim} is used in our optical flow loss to assess the similarity between the target image and warped image:
\begin{equation}
     \operatorname{SSIM}(x, y)=\frac{\left(2 \mu_x \mu_y+c_1\right)\left(2 \sigma_{x y}+c_2\right)}{\left(\mu_x^2+\mu_y^2+c_1\right)\left(\sigma_x^2+\sigma_y^2+c_2\right)},
\end{equation}
where $\mu_x$ and $\mu_y$ represent the mean pixel values of images $x$ and $y$, and $\sigma$ denotes the corresponding standard deviation. 

\paragraph{Census Transform Loss} We use ternary census transform loss~\cite{stein2004efficient,meister2018unflow,hafner2013census} as the second metric to evaluate optical flow performance: 
\begin{equation}
\mathbf{CT}\left(p, p^{\prime}\right)= 
\begin{cases} -1 & \text { if } p^{\prime}-p\geq \epsilon \\ 
+1 & \text { if } p - p^{\prime} \geq \epsilon \\
0 & \text { if } |p - p^{\prime}|<\epsilon
\end{cases}
\end{equation}
Given two input images, we compute the corresponding census-transformed images and compute the average difference between them as the loss.

\subsection{LiDAR to Range Image}
Since point cloud-based networks \cite{qi2017pointnet,qi2017pointnet++,wu2020pointpwc} are computationally demanding and complex to train, we convert the 3D scan points to sparse range maps with the help of the camera-LiDAR transformation by projecting 3D points onto the image plane with known camera intrinsics $\K$ and camera pose parameters $\mathbf{R}$ and $\mathbf{t}$:
\begin{equation}
\mathbf{K}[\mathbf{R} \mid \mathbf{t}]
=\left[\begin{array}{lll}
f & 0 & p_x \\
0 & f & p_y \\
0 & 0 & 1
\end{array}\right]\left[\begin{array}{llll}
r_1 & r_2 & r_3 & t_1 \\
r_4 & r_5 & r_6 & t_2 \\
r_7 & r_8 & r_9 & t_3
\end{array}\right].
\end{equation}
The 2D projection $\mathbf{p}=(u,v,1)^T$ of 3D point $\mathbf{P}=(x,y,z,1)^T$ is computed as 
\begin{equation}
    \mathbf{p} = \mathbf{K}[\mathbf{R} \mid \mathbf{t}]\mathbf{P}
\end{equation}
and rounded to the closest integer pixel coordinate.

\subsection{Runtime and Model Size} 
We report the runtimes and model sizes of RAFT, DeFlow-Cam, and DeFlow-Fusion in \cref{tab:runtime}.
Our camera-only baseline
is significantly smaller and faster,since we follow the lightweight design of PWC-Net~\cite{sun2018pwc} and Mono-SF~\cite{hur2020self}. Our fusion model has similar model size and runtime as RAFT. The increase in runtime and model size compared to the camera-only baseline  is caused by the additional depth encoder and the multi-level feature fusion, which in return offer marked performance gains. 
\begin{table}[h!]
\renewcommand{\arraystretch}{1.0}
\setlength{\tabcolsep}{4pt}
	\centering
 \resizebox{0.80\columnwidth}{!}{
	\begin{tabular}{l|c|c}
\toprule {Method}  & \multicolumn{1}{c}{Runtime [ms]} & \multicolumn{1}{|c}{\# params}\\
\midrule
{RAFT~\cite{teed2020raft}} &  171.3  & 5.26 M\\
{DeFlow-Cam (\textit{Ours})} & 54.1  & 4.16 M\\
{DeFlow-Fusion (\textit{Ours})} & 191.9  &  4.99 M\\

\bottomrule
\end{tabular}}
	\vspace{-1mm}
	\caption{Runtime and number of parameters for different models.}
 	\vspace{-2mm}
 	\label{tab:runtime}
\end{table}

\subsection{Additional Qualitative Results}
\begin{figure*}[h]
     \centering
        \includegraphics[width=1.0\linewidth]{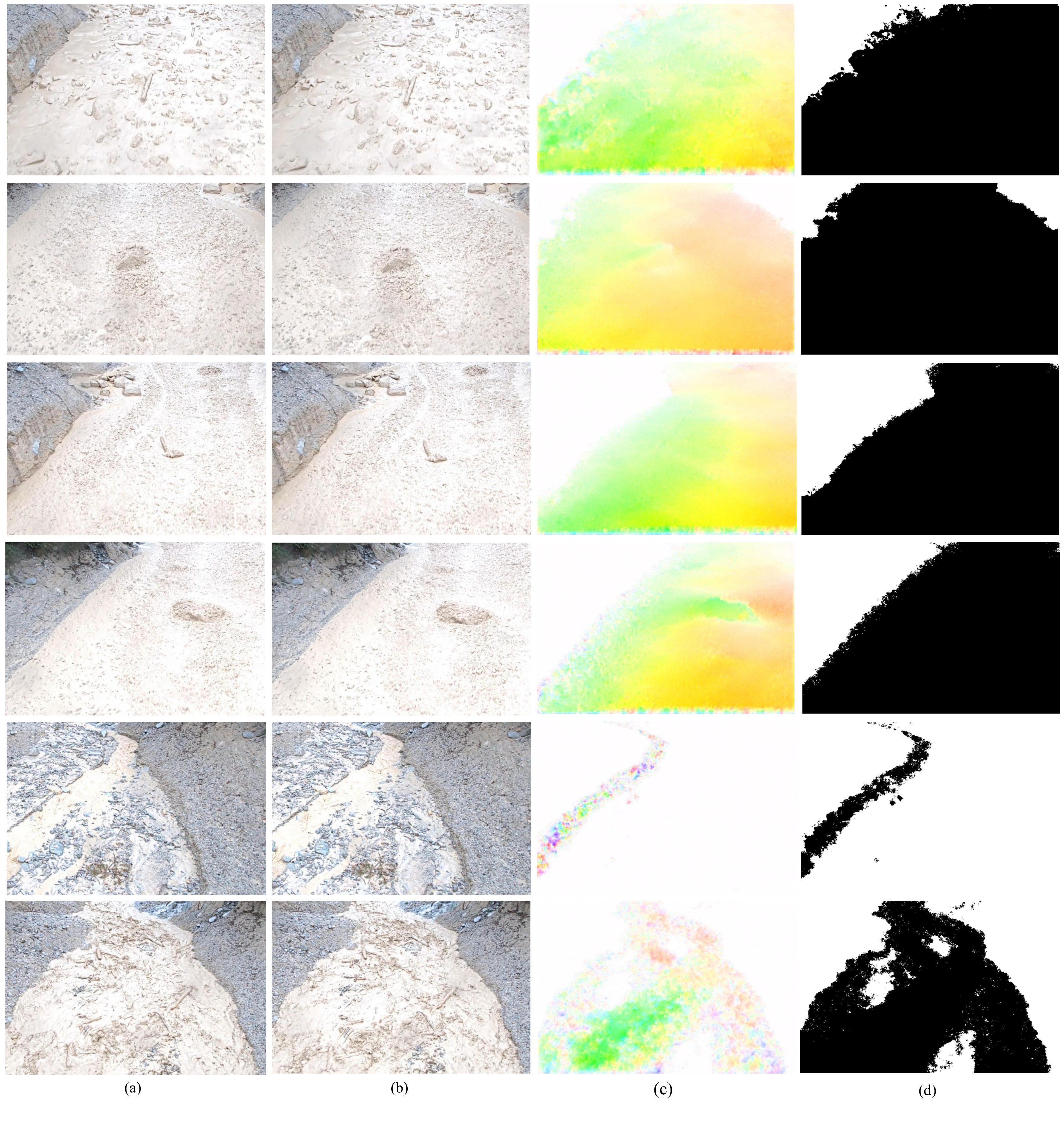}
   \vspace{-0.7cm}
       \caption{\textbf{Qualitative results} of optical flow (c) and binary (static/dynamic) segmentation (d). }
   \label{fig:add}
\end{figure*}

In \cref{fig:add}, we present additional visualizations of optical flow results, and of the static pixel masks used to enforce a static sensor pose.
To see the qualitative behaviour of our temporal smoothing module, \textit{readers are encouraged to watch the video in the supplementary material.}

\paragraph{Flow and depth trade-off}
We design our network to jointly estimate the depth and optical flow, which is a multi-task learning problem. We trade off the weights between $\lambda_{\text{flow}}$ and $\lambda_{\text{depth}}$ ($\lambda_{\text{flow}}+\lambda_{\text{depth}} = 1$), evaluate model performance accordingly and summarize the results in \cref{tab:ablation}. As expected, reducing $\lambda_{\text{flow}}$ results in a corresponding decline in the performance of the optical flow component. This is not surprising given that the training of self-supervised optical flow is complex and time-consuming, and the optical flow branch dominates the encoder and decoder, while the depth branch is trained more stably with direct supervision from LiDAR. For future applications of our approach, the trade-off between these two components can be determined based on the specific analysis requirements, such as surface motion or flow structure.
\begin{table}[h!]
    \setlength{\tabcolsep}{1pt}
    \renewcommand{\arraystretch}{1.3}
    \centering	
    \resizebox{1.0\columnwidth}{!}{
    \begin{tabular}{cc|cc|ccccc}
    \toprule
      \multirow{2}{*}{$\lambda_{\mathrm{flow}}$}& \multirow{2}{*}{$\lambda_{\mathrm{depth}}$} &  \multicolumn{2}{c|}{Optical Flow$\downarrow$} &   \multicolumn{4}{c}{Depth Estimation$\downarrow$} \\
    & & Census Loss  & RMSD & $\mathrm{MAE_{10}}[m]$ & $\mathrm{MAE_{30}}[m]$ & $\mathrm{MAE_{50}}[m]$  & $\mathrm{Abs.\: Rel. [\%]}$  \\
    \midrule
    0.9& 0.1& \textbf{0.104} &  \textbf{6.83} & 0.046 & 0.130 & 0.176 &0.79  &\\
    0.8& 0.2& 0.106 &  7.03 & 0.042 & 0.118 & 0.162 &0.72  &\\
    0.7& 0.3& 0.108 &  7.10 & 0.042 & 0.121 & 0.164 &0.74  &\\
    0.6& 0.4& 0.107 &  7.19 & 0.040 & 0.136 & 0.175 &0.79  &\\
    0.5& 0.5& 0.111 &  7.41 & 0.050 & 0.115 & 0.143 &0.75  &\\
    0.4& 0.6& 0.111 &  7.43 & \textbf{0.034 }& 0.150 & 0.196 &0.82  &\\
    0.3& 0.7& 0.119 &  8.06 & 0.052 & 0.131 & 0.166 &0.81  &\\
    0.2& 0.8& 0.117 &  7.90 & 0.053 & 0.178 & 0.231 &1.02  &\\
    0.1& 0.9& 0.138 &  9.51 & 0.039 & \textbf{0.097 }& \textbf{0.137 }&\textbf{0.61 } &\\
  
    \bottomrule
    \end{tabular}
    } 
    \vspace{-1mm}
    \caption{\textbf{Ablation study on optical flow and depth trade-off.}} 
    \vspace{-5mm}
    \label{tab:ablation}
\end{table}
\end{document}